\documentclass[letterpaper, 10pt, conference]{cssconf}
\IEEEoverridecommandlockouts                              % This command is only
\usepackage{graphicx} % for pdf, bitmapped graphics files
\usepackage{xspace}
\usepackage{url}
\usepackage[usenames,dvipsnames]{color}
\usepackage{bm}

\newcommand{\acronym}[1]{\textsc{#1}\xspace}
\newcommand{\acronymnospace}[1]{\textsc{#1}\xspace{}}%force not to put a space after #1
\newcommand{\virgo}[1]{``#1''}
\newcommand{\ie}{i.e.\@\ }% 'i.e.' *always* has a blank space after it
\newcommand{\eg}{e.g.\@\ }% 'e.g.' *always* has a blank space after it
\newcommand{\hyq}{{HyQ}\xspace}
\newcommand{\SL}{\acronym{sl}}
\newcommand{\uml}{\acronym{uml}}
\newcommand{\xml}{\acronym{xml}}
\newcommand{\dof}{\textsc{d}o\textsc{f}\xspace}
\newcommand{\dofs}{\textsc{d}o\textsc{f}s\xspace}
\newcommand{\tmx}[3]{\phantom{}_#2 #1_#3}
\newcommand{\tmxX}[2]{\tmx{X}{#1}{#2}}
\newcommand{\comment}[2]{}%\textcolor{#1}{\large \textbf{\textit{[#2]}}}}%make sure no comments left in the paper
\newcommand{\mfnote}[1]{\comment{OliveGreen}{MF: #1}}

\newcommand{\textcode}[1]{\texttt{#1}}
\newcommand{\dsl}{\acronym{dsl}}
\newcommand{\dsls}{\acronymnospace{dsl}s\xspace}
\newcommand{\myheader}[1]{}%\textbf{#1}\\}
\newcommand{\matlab}{\textsc{matlab}\xspace}

\title{\LARGE \bf A Domain Specific Language for kinematic models and
fast implementations of robot dynamics algorithms}

\author{Marco Frigerio, Jonas Buchli and Darwin G. Caldwell% <-this % stops a space
\thanks{Marco Frigerio, Jonas Buchli and Darwin G. Caldwell are with the
        Department of Advanced Robotics, Istituto Italiano di Tecnologia, via
        Morego, 30, 16163 Genova.
        {\small\{\texttt{marco.frigerio},
                 \texttt{jonas.buchli},
                 \texttt{darwin.caldwell}\}
                      \texttt{@iit.it}
                      }%
}}

\begin{document}
%%%%%%%%%%%%%%%%%%%%%%%%%%%%%%%%%%%%%%%%%%%%%%%%%%%%%%%%%%%%%%%%%%%%%%%%%%%%%%%%
%
\maketitle
\thispagestyle{empty}
\pagestyle{empty}
%
%%%%%%%%%%%%%%%%%%%%%%%%%%%%%%%%%%%%%%%%%%%%%%%%%%%%%%%%%%%%%%%%%%%%%%%%%%%%%%%%
\begin{abstract}
Rigid body dynamics algorithms play a crucial role in several components of a
robot controller and simulations. Real time constraints in high frequency
control loops and time requirements of specific applications demand these
functions to be very efficient. Despite the availability of established
algorithms, their efficient implementation for a specific robot still is a
tedious and error-prone task. However, these components are simply necessary
to get high performance controllers.\\
To achieve efficient yet well maintainable implementations of dynamics algorithms
we propose to use a domain specific language to describe the
kinematics/dynamics model of a robot. Since the algorithms are parameterized on
this model, executable code tailored for a specific robot can be
generated, thanks to the facilities available for \dsls.
This approach allows the users to deal only with the high level description
of their robot and relieves them from problematic
hand-crafted development; resources and efforts can then be focused on open
research questions.\\
Preliminary results about the generation of efficient code for inverse dynamics
will be presented as a proof of concept of this approach.
\end{abstract}
%
%%%%%%%%%%%%%%%%%%%%%%%%%%%%%%%%%%%%%%%%%%%%%%%%%%%%%%%%%%%%%%%%%%%%%%%%%%%%%%%%
\section{INTRODUCTION}\label{sec:intro}
%%%%%%%%%%%%%%%%%%%%%%%%%%%%%%%%%%%%%%%%%%%%%%%%%%%%%%%%%%%%%%%%%%%%%%%%%%%%%%%%
\myheader{Intro, software for robotics is crap}
According to the presentation of the joint research project \acronym{brics},
aiming at identifying best practices in the development of robotics
systems, such development process often lacks of a rigorous structure and
principles \cite{bischoff:2010:brics}, even after
decades of research in the field. A typical example is software development for
robotics, where the lack of design and identification of effective
abstractions lead to the development of code--driven systems as opposed to
model--based ones. In this regard, in \cite{schlegel:2009:model_driven} the
authors point out the gap between the experience available in robotics and
the exploitation of such knowledge  for a proper software development process.

For the robotics research community as well as for a widespread adoption of
robotic technology it is central to have flexible yet reliable software: a
typical academic research unit can not afford the same resources to develop
reliable software as an airplane or a car manufacturer, yet requires
dependable and flexible software for similar complex machinery, in order to
address open research questions.\\
Developing software for robots is among the most demanding and
complex software engineering challenges due to a list of strict and
partially conflicting requirements and the sheer complexity arising
from the many tasks such a software has to perform in a well orchestrated
manner.

More specifically, typical requirements for robot controller software are:
%
%\myheader{Common requirements for robots software}
%
\begin{itemize}
\item Real time capability: specific sections of the program must be able
      to run in a hard real time context (\eg a 1KHz force control loop).
\item Safety guarantees: a high level of robustness of the whole system is
      desired (\eg if dealing with a potentially dangerous robot, for people or
      for itself).
\item Generation and deployment of components for multiple targets
      (\eg different programming languages or different hardware platforms).
\item Integration of many different resources (sensors, motors,
  processors etc), with different physical interfaces and APIs.
\item Varieties of time constants and resource requirements: robotics
      applications must integrate components that need to run at different
      frequencies, with
      diverse usage of computation and memory resources (\eg
      the sampling of a fast analog sensor and a stereo camera).
\end{itemize}
Satisfying such requirements translates in many software engineering challenges
\eg concerning also the architectural level of
design \cite{garlan:1993:intro}:
\begin{itemize}
\item Domain models: finding appropriate abstractions for very common components
      and \emph{recurring problems} in robotics, to establish best practices and
      principled, general solutions (\eg a reference C implementation of a PD
      controller or a general model of virtual components for operational space
      control \cite{pratt:2001:virtual}).
\item Clear separation between control logic and task logic: it is desirable to
      be able to run exactly the same task code both against a simulator and on
      the real robot.
\item Flexible yet resource efficient and real time capable memory
  management.
\item Automatic unit testing.
\item Tools to assess memory and time complexity.
\item Integration of controller code: strategies to include components designed
      with different tools, such as \matlab and \textsc{simulink}
      \cite{matlab}.
\item Automatic generation of infrastructure code: \eg common components of
      simulators, coordinate transform matrices. It is desirable to avoid
      error-prone development by hand if this can be automated according to
      established models.
\item Logging/debugging facilities: proper
      diagnostic tools that also satisfy the other requirements (\eg a
      real time compatible logger).
\item Graphical interfaces: visualization of the robot, the state of
      its controllers, the layout of reference frames and so on; this
      dramatically reduces the effort for debugging.
\end{itemize}
%%%%%%%%%%%%%%%%%%%%%%%%%%%%%%%%%%%%%%%
\subsection{Contribution and motivation}
In this paper we address the automatic code generation of rigid-body
dynamics algorithms for simulation and control of a robot under real time
constraints, based on a general domain model.
We will focus on robots assembled in chains or branched chains of rigid links.\\
Code used for model based controllers is a typical example of software with an
apparent trade-off between flexibility/maintainability and efficiency. On one
hand a rigid body model is a generic description of a robot that
naturally lends itself for a rather general implementation, \eg with object
oriented code. On the other hand it is critical that such code does not violate
real time constraints (\eg by system calls such as those for dynamic memory
allocation
or file access) and is ideally running as fast as possible (\eg exhaustive
evaluations in sampling based planner algorithms or fast control loops).

To address and resolve this apparent trade-off, we propose a simple
meta--model for the generalization of kinematics/dynamics models of
robots, a Domain Specific Language (\dsl)
for specifying conveniently such models, and a transformation step
built around the \dsl for the generation of optimized rigid body
dynamics algorithms.\\
The basic idea comes simply from the observation that dynamics algorithms
are general and parametrized on the kinematic description of a robot --
often called the robot \emph{model} \cite{featherstone:2008:rbda} --
which is relatively compact and based on a common schema, but fully
specifies the physics of the system.
Thus it is sensible to look for a high level representation of the robot
models, which can be easily constructed by hand, while exploiting automated
procedures to turn such information into executable procedures tailored for
the specific robots.
As a an example, we will address a real time capable C++ implementation
of the recursive Newton-Euler algorithm.

This approach is not new and with some variations it is adopted by simulators
and software packages commercially available. For
instance, SD/FAST \cite{web:sdfast} is a complex and rich simulator of mechanical
systems that produces C or Fortran implementation of the equations of motions
for the given system. Similarly, Robotran \cite{web:robotran} targets multi-body
dynamics applications; after reading a user model defined with a graphical
editor, it can output symbolic equations of motion and perform simulations
interacting with \matlab.\\
\SL is a rigid body dynamics simulator and robot controller package -- we are
currently using it for our research --, which uses its own description of
kinematics and can produce highly optimized C code \cite{schaal:2006:sl}.
The performance of such code is definitely high, but quite
some improvement can be introduced with regard to the flexibility and
usability of the generation process.
\mfnote{Move paragraph in Related Works?}

Our aim is to collect existing strategies and provide a coherent approach
based on a sound model and dependent solely on open source technologies.
We aim at tools that possibly target different robotics platforms, which
can be easily adopted by users in this field and help improving the quality
of their work.\\
As a matter of fact, even though the modeling of kinematic trees and the
related algorithms for kinematics and dynamics are well known in the robotics
community and have
been extensively studied in the past decades, they still represent an obstacle
for the development of new robotics applications: a lot of initial development
that targets such issues is required to make any robot operational, it is critical
for the control and simulation but it is often not the focus of the research per
se (for example if one wants to test his own learning algorithms on a new
manipulator). Thus, researchers would benefit from an automatic
implementation in terms of robustness of the code and time spent during the
start-up of the project.

A general dynamics library (such as \acronym{ode} \cite{ode.org}) -- which
would necessarily require the robot model as a \emph{parameter} -- could solve this
problem. But the point of being able to generate a specific \emph{instance} of
the algorithm is efficiency without loosing flexibility. With our approach, one can target
different platforms and apply custom optimizations. This aspect is further discussed
in Section \ref{sec:code}.
%%%%%%%%%%%%%%%%%%%%%%%%%%%%%%%%%%%%%%
\subsection{Domain Specific Languages}
A Domain Specific Language (\dsl) is a formal language suitable to
represent some sort of specification related to a precise class of problems
only \cite{mernik:2005:dsl}. The syntax and the semantic of the language are
explicitly designed to have a limited expressiveness in general, which is paid
off by the ease with which the elements of the target domain can be
represented.\\
Moreover, a \dsl itself can be implemented -- by specifying the custom syntax,
providing a parser and further required facilities -- with a technology independent from the final
platform, \ie the target for eventual executable artifacts. So for instance
one can have multiple code generators that take an instance document of the
\dsl (\eg a plain text file) and output code in different languages.

Although quite general, this description already suggests how the \dsl
technology nicely fits our problem and requirements, and therefore
it is sensible to adopt it for our purposes: first, robot models can be
described by easy-to-read text files which follow a custom syntax tailored
to the specific domain. In the Robot Operating System \acronym{ros}
\cite{quigley:2009:ROS}, for instance, model files need to be provided
as \xml which is harder to read and maintain; in the OpenHRP simulator \cite{web:openhrp},
the language for the models comes from the 3D modeling field, and mixes
graphical aspects and sensors with kinematics parameters.\\
Then, these documents can be parsed, checked and subject to custom transformations
like generating code. The \dsl allows to nicely decouple the simple model that
needs to be built by the user, and the coding, which is more complex and can be
partially automated.

The rest of the paper is organized as follows: Section \ref{sec:code} describes
a code generator which exploits a priori knowledge of the robot; Section
\ref{sec:model} presents the general structure of kinematic models while
Section \ref{sec:dsl} describes a \dsl to provide robot specific descriptions.
Finally, section \ref{sec:related} presents some related work and section
\ref{sec:end} discusses improvements and future developments.
%
%%%%%%%%%%%%%%%%%%%%%%%%%%%%%%%%%%%%%%%%%%%%%%%%%%%%%%%%%%%%%%%%%%%%%%%%%%%%%%%%
\section{EFFICIENT CODE GENERATION}\label{sec:code}
%%%%%%%%%%%%%%%%%%%%%%%%%%%%%%%%%%%%%%%%%%%%%%%%%%%%%%%%%%%%%%%%%%%%%%%%%%%%%%%%
Rigid body dynamics algorithms can be used in a number of components
of the software system of a robot: model based control (e.g. impedance
control, inverse dynamics), simulation (e.g. physics based
simulation), planning (e.g. kino-dynamic planning). In some
applications (e.g. simulation) minimum time of execution is desired,
while in other applications (real time planning and control) a certain
maximum time of execution of the code is a strict requirement.\\
On the other hand, manual coding of these routines is a non-trivial
and error-prone task, and the demand for optimizing the execution time
only makes the task harder. Therefore, leaving it to a computer and
concentrate on higher level aspects of the research question, whenever
possible, is an effective approach.

Other arguments for efficient implementations include the persistence
in robotics of constraints due to space or power availability. Often
one must adopt embedded computers, less powerful than regular desktop
machines.\\
A user
might also be simply interested in having \emph{full} control on the
software of the robot, and would therefore appreciate to develop (once
for all) his own code generator without external dependencies. This
requirement might arise when dealing with low level, hard real time
code for a machine that requires strict control to guarantee
safe operation (such as the hydraulic quadruped robot we are developing
at our lab, \hyq \cite{semini:2011:hyqjournal}).

In this paper we focus on the Newton-Euler inverse dynamics algorithm as the
reference example (see \cite{featherstone:2008:rbda, featherstone:2010:tut1,
featherstone:2010:tut2} for detailed explanation of the algorithm).
The purpose of inverse dynamics is computing the following function for
multi body systems:
\begin{equation}
\bm{\tau} = f(\ddot{\bm{q}},~\bm{q},\dot{\bm{q}})
\end{equation}
where $\bm{q}$ and $\dot{\bm{q}}$ are the \emph{actual} joint position
and velocity vectors, $\ddot{\bm{q}}$ is the \emph{desired} joint
acceleration vector, and $\bm{\tau}$ contains the forces required to
achieve such accelerations.\\
As pointed out in \cite{featherstone:2008:rbda}, an additional, implicit
input is the system model for which forces have to be computed.
Exploiting prior knowledge about the structure and the
parameters of the robot, we can resolve that dependency but also generate
optimized code, by avoiding any logic that deals with a generic case
(\eg loops) and especially by exploiting numerical
properties (\eg avoiding multiplication with
zero to simplify matrix operations). For instance, in the assumption of
having only plain prismatic or revolute joints, the
matrix $S$ describing the motion subspace of a joint is a single column
vector with only one non-zero element,
thus operations involving this matrix can be greatly simplified.

Another well known advantage of code generation based on a \dsl is the
possibility to target different languages and platforms. For instance,
even if the purpose of \matlab is certainly not to achieve top
speed, one would still benefit from optimized (if automatically generated)
algorithms for simulations and rapid prototyping of algorithms.
%%%%%%%%%%%%%%%%%%%%%%%%%%%%%%%%%%%%%%%%%%%%%%%%%%%%%%%%%%%%%%%%%%%%%%%%%%%%%%%%
\section{MODELING KINEMATIC TREES}\label{sec:model}
\subsection{Introduction}
In this paper we deal with robot \emph{models} -- descriptions implying a
certain degree of abstraction -- related to kinematics and dynamics.
The main assumption underlying
these models is that all the bodies comprising the system are perfectly rigid.
From the dynamics point of view (\ie rigid body dynamics), the basic model also
assumes idealized sources of \emph{generalized} force (\ie force/torque) that
move the bodies; the information required to compute the effect of forces
is given by
the inertia parameters of the bodies, \ie mass, position of the center of mass,
inertia matrix.\\
Concerning kinematics, we shall give here a brief description of
the structure of the models and the amount of information they embed, to
provide the background for the rest of the paper. For an extensive and
authoritative treatise on these topics, see \cite{siciliano:2009:robo_book,
featherstone:2008:rbda}.

In kinematic models, a robot is an assembly of links and joints: a link is a
rigid body with inertia properties while a joint represent a \emph{constraint}
between exactly two bodies (the predecessor and successor), which would otherwise
be fully free to move relatively to each other.
Such a constraint is not purely a rigid junction since the joint guarantees
certain \emph{degrees of freedom} (\dof) to the attached link. A specification
of the nature of each joint is obviously required.\\
The description of the whole structure of a robot is topological, that is, it
can be simply represented by a graph where joints are arcs and bodies are nodes
(quite the contrary of what graphical intuition might suggest).
For simplicity, we will focus only on kinematic trees (\ie no loops in the
structure), which represent a wide class of the robots used in industry and
research; the full generalization of the model is one of the natural topics
for future development, and can be done by integrating the
methods described in \cite{featherstone:2008:rbda} in our \dsl framework.

\emph{Reference frames:}
The geometry of the bodies and their connections is required to dynamically
compute the \emph{pose} of the bodies, the dynamical effects of the movements,
such as Coriolis and centrifugal forces, and so on. To this end, various
reference frames must be placed in known points
of every body and every joint of the tree. The parameters for a set of
\emph{transformations} among different frames \emph{plus} a convention about
the placement of them (\eg the $z$ axis of a joint reference is always
aligned with the rotation axis) basically encode all the required information.

%%%%%%%%%%%%%%%%%%%
\begin{figure}[tbp]
	\centering
	\includegraphics[width=0.9\columnwidth]{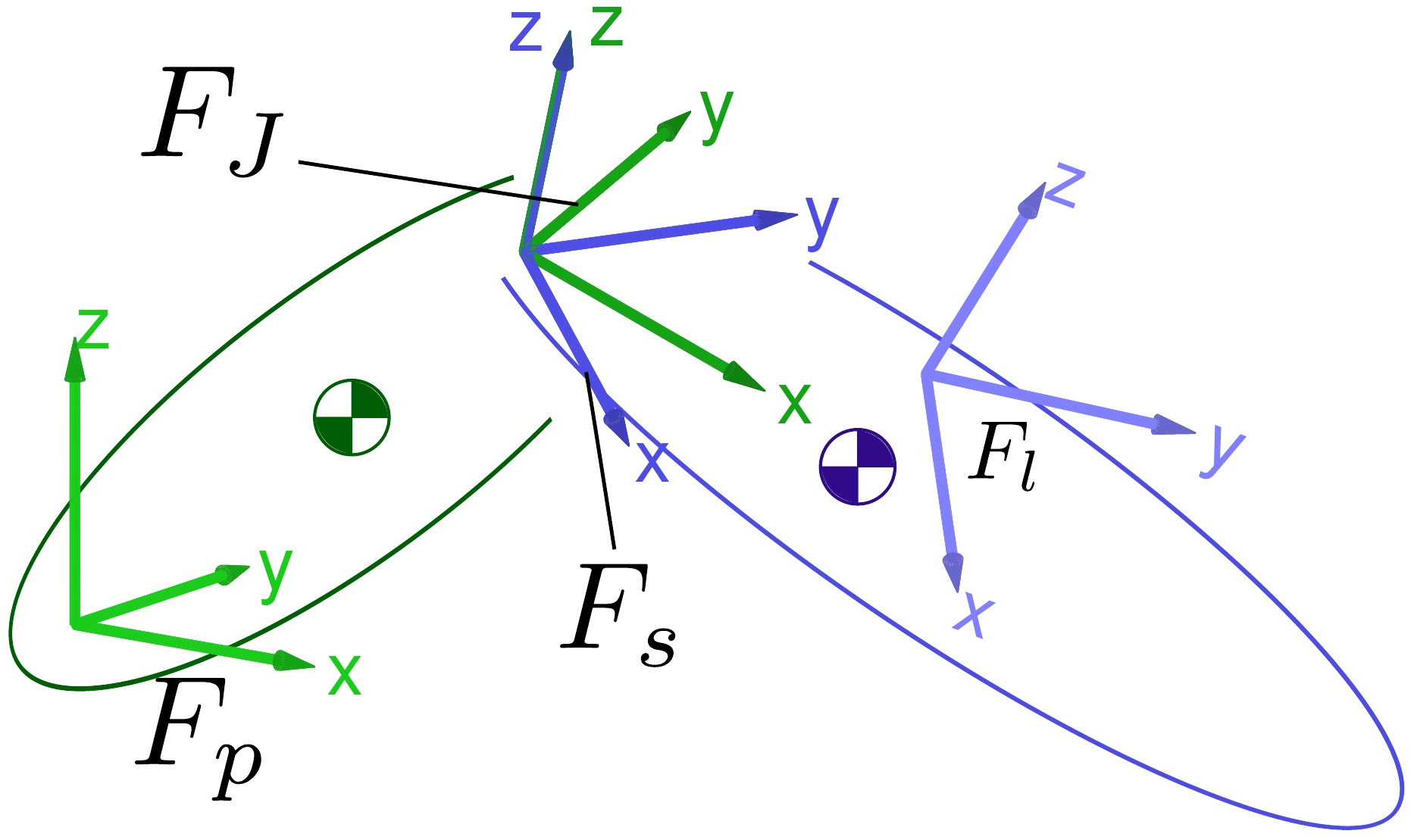}
	\caption[Layout of reference frames for a generic section of a
	kinematic chain]
	{Layout of reference frames for a generic section of a
	kinematic chain. $F_p$ and $F_s$ are the frames respectively of
	the predecessor and successor link of joint $J$, whose frame is
	$F_J$. $F_p$ and $F_J$ do not move with respect to each other, while
	$F_J$ and $F_s$ do, according to the joint behavior.
	$F_l$ shows a possible additional frame located on the link.}
	\label{fig:frames}
\end{figure}
%%%%%%%%%%%%%
Figure \ref{fig:frames} shows the layout of frames in a general case. For more
information about the convention please refer to \cite{featherstone:2008:rbda}
and \cite{featherstone:2010:tut2}; in the following we state only some
observations relevant for the development of our \dsl.\\
We emphasize that the transform $\tmxX{J}{p}$ for the joint frame
is a \emph{constant}, since it describes the placement of the joint
expressed in the reference of the predecessor link (\ie $\tmxX{J}{p}$
depends on static, geometrical parameters of the robot).\\
Furthermore, we note that for each joint there are two frames, which coincide
when the joint status (\ie the actual angle or displacement) is zero.
Only the second frame moves as the joint moves, since it
is attached with the successor link. As in \cite{featherstone:2008:rbda}
this frame ($F_s$) is chosen to be the reference for this link. Among the other
things, this implies:
\begin{itemize}
 \item no transformation parameters have to be associated with the link, since
 its frame is completely determined by the convention and the joint status.
 \item The generic transform $\tmxX{s}{J}$ between the two frames on the joint
 ($F_J$ and $F_s$) is the only
 one which depends on the joint status. Note that $\tmxX{s}{J}$ captures the
 \emph{type} of the joint as well (\ie rotational or prismatic).
% \item The translation encoded in $X_J$ corresponds to the distance between the
% axes of two joints attached to the link \draftnote{remove this?}
\end{itemize}

Even if adopting this convention, for further flexibility another frame might
be added anywhere on the link, according to any user preference or requirement
(\eg to express more conveniently the position of certain sensors placed on the
link).
%However, in general it might not be a good idea to use it in place of
%$F_s$, for efficiency reasons; for instance
%
%\draftnote{the minimum amount of information to model .. is as follows..}
%
%%%%%%%%%%%%%%%%%%%%%%%
\subsection{\uml model}\label{sec:uml}
%%%%%%%%%%%%%%%%%%%%
\begin{figure*}[tbp]
	\centering
	\includegraphics[width=0.75\textwidth]{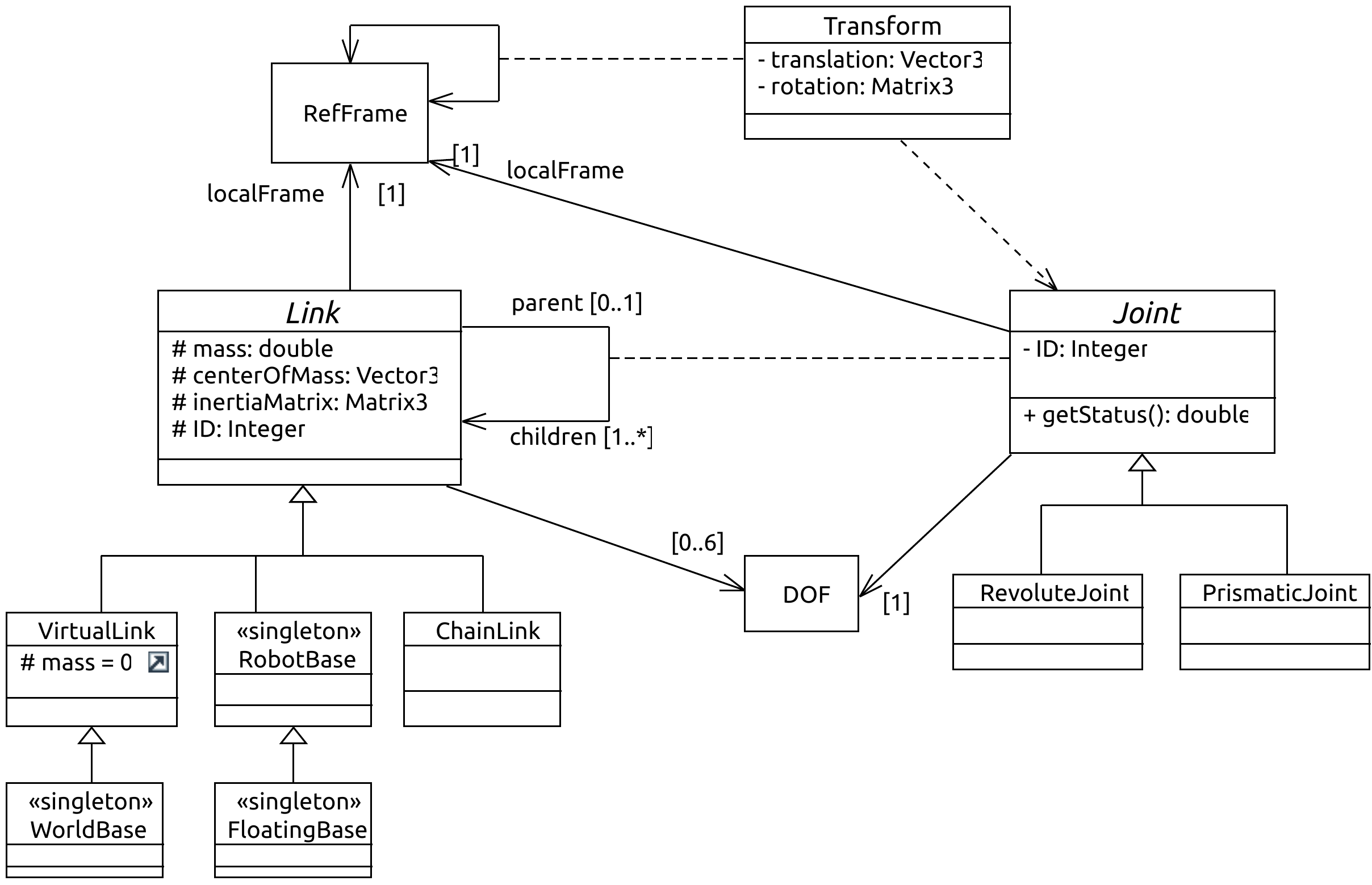}
	\caption{The kinematic tree (meta)model as an \uml class diagram.}
	\label{fig:uml_model}
\end{figure*}
%%%%%%%%%%%%%
Figure \ref{fig:uml_model} shows an \uml class diagram representing the key
elements described before. The diagram is simple but general, and can be
applied to almost any robot made by rigid links.\\
The central classes, quite intuitively, are \textcode{Link} and
\textcode{Joint}.
Joints induce a parent--child relationship among links, which is characterized
by the type of joint. We chose to model this relationship by making \textcode{Joint} an
association class connected to the self--association for \textcode{Link}.
To keep the model simple we consider only 1-\dof joints, since actual composite
joints (as a three \dofs ball joint)
can be represented by primitive ones connected by virtual dimensionless
links (irrelevant for kinematics and dynamics computations -- see below).\\
The association between \textcode{Joint} and the class \textcode{DoF} basically
models the intrinsic property that tells which relative movement is allowed by
the joint. Links, on the other hand, have certain degrees of freedom as a
consequence of the kinematic configuration, and they have a similar association
as well.

Any link can have multiple children, which corresponds to a branched structure of
the robot. On the other hand, as mentioned before we do not consider loops, which
would require another type of joint (\ie a loop joint) which does not determine
any new child link but rather connects two existing links. The abstract class
\textcode{Link} actually models any rigid body, and has a few subclasses to
differentiate particular cases:
\begin{itemize}
\item \textcode{ChainLink}: a generic piece of the kinematic chain, what is
      usually referred to as link;
\item \textcode{RobotBase}: a special link which represents the
      \virgo{root} of the kinematic tree. Can be floating if the robot is a
      mobile one. Note the stereotype \textcode{Singleton}, since there is
      only one base for each robot;
\item \textcode{VirtualLink}: a dimensionless body to allow the representation
      of complex joints (see above); this class explicitly forces the inertia
      parameters of its instances to be zero. Floating base robots
      can be thought as connected via a virtual six-\dofs
      joint (\ie no constraint), to an arbitrary point in the world, which
      is a virtual link as well: \textcode{WorldBase} (cf.
      \cite{featherstone:2008:rbda}).
\end{itemize}

Finally, the conceptual model of Figure \ref{fig:uml_model} describes reference
frames through the class \textcode{RefFrame} and the associations with
\textcode{Link} and \textcode{Joint} (see section \ref{sec:model}); however,
since a frame per se does not really have any property (we assume only
right--handed coordinate systems) or behavior, we observe that the relevant
information is instead in \textcode{Transform}, which provides the transformation
parameters for a given couple of frames.
%%%%%%%%%%%%%%%%%%%%%%%%%%%%%%%%%%%%%%%%%%%%%%%%%%%%%%%%%%%%%%%%%%%%%%%%%%%%%%%%
\section{THE DSL}\label{sec:dsl}
\dsls can be roughly divided into two categories, internal and external, the
former being built through a particular usage of an existing language, while
the latter is independent and usually has a custom syntax
\cite{fowler:2010:dsl}. We shall choose an external \dsl, whose
model documents can be plain text files, with a clear aspect (syntax) and
intuitive semantics.

As argued in \cite{fowler:2010:dsl}, a proper \dsl design would not be
complete without an underlying domain model, for which \dsl documents
are just a specification of its instances%
\footnote{Actually Fowler uses the term \virgo{semantic model},
to mean a part of the whole domain model, and identifies each \dsl document with
\emph{a} semantic model, rather than talking about instances.}%
.
Our domain model is described in Section \ref{sec:model} and its instances are
specific robot models (so we can refer to the former as the \emph{meta}--model);
each \dsl document has to carry the information to populate one of such models
\ie telling the number of links/joints, their type, their attributes, and so on.
The grammar, which must specify the structure of such documents, is naturally
inspired by the meta--model, which defines the structure of the
\emph{information} carried in the documents. Therefore, after the model had been
established reasonably, the design of the grammar was quite straightforward.
The required effort was limited and subject to a confident understanding
of the domain.

\begin{figure}[tbp]
	\centering
	\includegraphics[width=0.8\columnwidth]{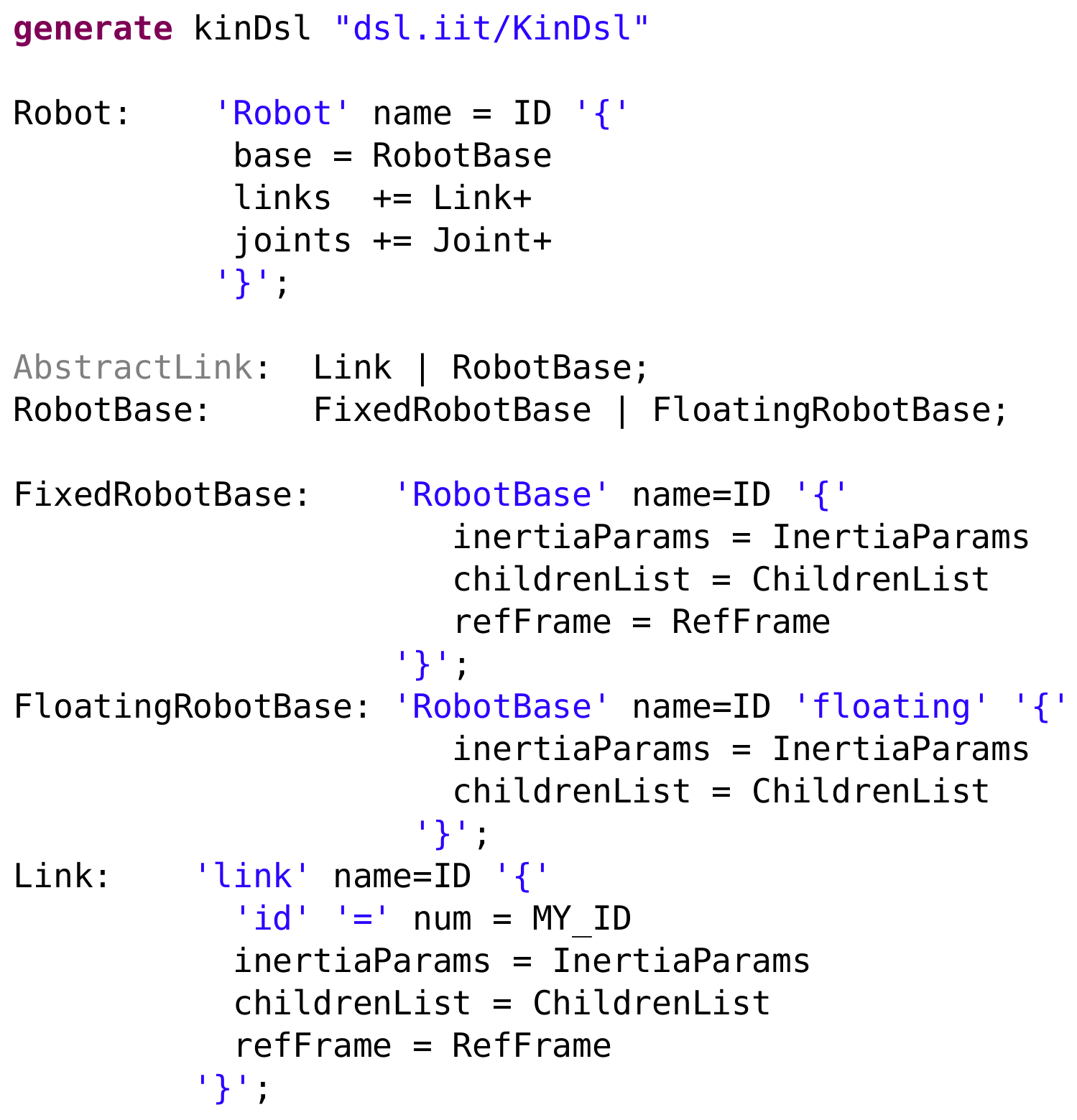}
	\caption[An excerpt of the \dsl grammar]
	{An excerpt of the \dsl grammar designed with Xtext.
	\textcode{Inertia\-Params} defines how to write in the document
	the inertia parameters of the bodies. \textcode{ChildrenList}
	allows to insert a list of links, while \textcode{RefFrame}
	takes care of the rotation and translation parameters of a
	coordinate transform. See also Figure \ref{fig:dslhyq}.}
	\label{fig:grammar}
\end{figure}

Obviously the grammar of the \dsl also provides additional syntax elements
to improve readability. See Figure \ref{fig:grammar} for an excerpt.
%%%%%%%%%%%%%%%%%%%
%%%%%%%%%%%%%
%
\subsection{Tools}
The cost for the syntactic freedom associated with an external \dsl is the
need to develop a custom grammar and
the associated parser, but luckily there are effective tools to support
these activities. We have adopted Xtext, a framework based on the Eclipse
platform that supports the creation of a complete language
infrastructure \cite{eysholdt:2010:xtext}; both software packages are open source tools.
In particular, Eclipse is a rich development environment widely used in
different domains, equipped with large support for model--driven development
\cite{gronback:2009:eclipse_modeling} and adopted in the robotics community
as well \cite{bischoff:2010:brics}. Xpand/Xtend are the related languages
to specify the templates for text generation.

However, it is important to note that Xtext/Eclipse can output a stand-alone
package containing the main tools related to the \dsl (\ie the parser and the
code generator), which can then be distributed and used independently of
it.
The only requirement is a Java interpreter, which is a widely
adopted technology. Eclipse/Xtext provides also rich text editing features to
write \dsl documents, but any plain text editor can be used.
\subsection{Example: a quadruped robot}
%%%%%%%%%%%%%%%%%%%
\begin{figure}[tbp]
	\centering
	\includegraphics[width=0.8\columnwidth]{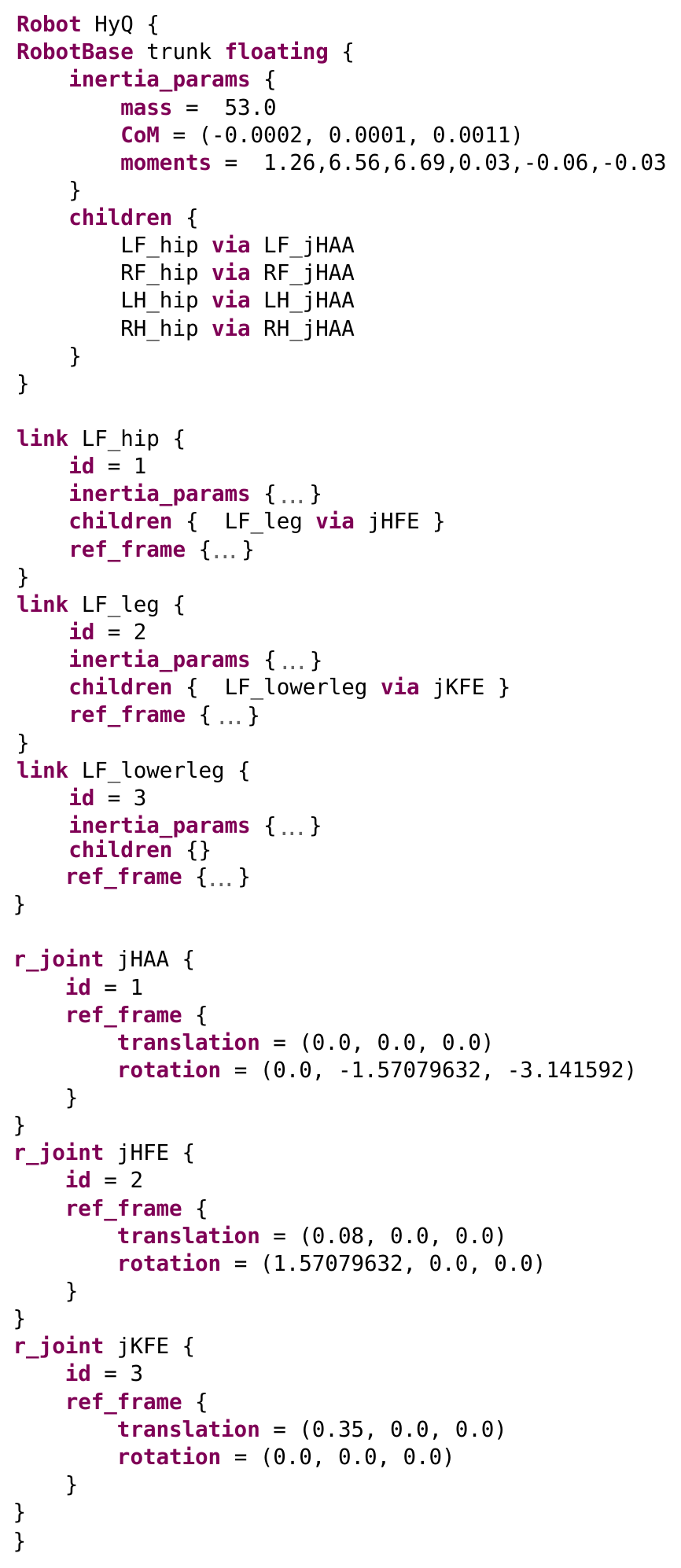}
	\caption[An excerpt of the \dsl document for \hyq]
	{An excerpt of the \dsl instance document modeling the quadruped
	robot \hyq. It shows the trunk and the parts of one leg; the other legs are
	almost identical. LF stands for left-front, RH for right-hind, and so on;
	jHAA is joint-Hip-Abduction-Adduction, jHFE is for Hip-Flexion-Extension,
	jKFE for Knee-Flexion-Extension. Inertia parameters and reference frames for the leg links
	are hidden to keep the image small.}
	\label{fig:dslhyq}
\end{figure}
%%%%%%%%%%%%%
To provide further insight on the structure of our \dsl, Figure \ref{fig:dslhyq}
shows a section of the description of our mobile, four legged robot \hyq
\cite{semini:2011:hyqjournal} as an example. \hyq has a trunk (the floating
base) and four identical legs -- left front, right front, left hind and right
hind -- each composed of three links: hip, upper leg and lower leg. As you can
see from the listings, the floating base does not specify any reference frame
transform, since there are no constraints between the world and the body and
all the parameters of such transform are free.
%%%%%%%%%%%%%%%%%%%%%%%%%%%%%%%%%%%%%%%%%%%%%%%%%%%%%%%%%%%%%%%%%%%%%%%%%%%%%%%%
\section{EXPERIMENTS AND RESULTS}
Once the \dsl is completed, creating new robot descriptions is a
matter of minutes, since the \dsl is simple and intuitive (most of the
time is spent looking in the robot documentation for the inertia
parameters and the frame transformations). If the code generator is properly
verified, then it is impossible to introduce low level bugs such as
memory leaks in this step.\\
For a
proof of concept of the proposed approach we chose the C++ language
and the Eigen library for linear algebra \cite{web:eigen}; this
allowed to have more compact and readable code, so that it is easier
to debug during the first experiments.  Eigen is a modern, carefully
designed and quite well documented library for efficient computations
with matrices, adopted for instance in \acronym{ros}
\cite{quigley:2009:ROS}.

However, whether using an external library is appropriate given the discussed requirements, is
something we have to establish with experimentation: on one hand these
libraries provide very efficient optimization (\eg avoiding
temporaries and exploiting sparsity), and also allow to have clearer
code. The other solution (mandatory if similar facilities are not
available but speed is still of
concern) is to manually address each single operation of the algorithm producing
low level, basic instructions only when necessary; this is the most
inconvenient approach, results in not so clear templates and code but
it is the most efficient. In addition, adopting a library injects an
external dependency and might not be trivial to use it properly (in
our opinion, an effective usage of Eigen requires some experience,
due for instance to the complexity of expression templates).
%%or very specific design decision the user needs to be aware of).

For numerical correctness, we have tested our implementation of the
inverse dynamics algorithm against the \matlab code available
on Featherstone's web page \cite{web:roy}, comparing the numerical
output for different robot models and different inputs ($\bm{q}$,
$\dot{\bm{q}}$ and $\ddot{\bm{q}}$).\\
As far as performance is concerned, instead, we made some comparisons
with the \SL simulator;
as mentioned in Section \ref{sec:intro}, this software generates a
highly optimized, low level C code implementation, whose performance
can very well be considered as a reference. \SL adopts exactly the
inconvenient-but-fast approach described before.

%%%%%%%%%%%%%%%%%%%
\begin{figure}[tbp]
	\centering
	\includegraphics[width=0.7\columnwidth]{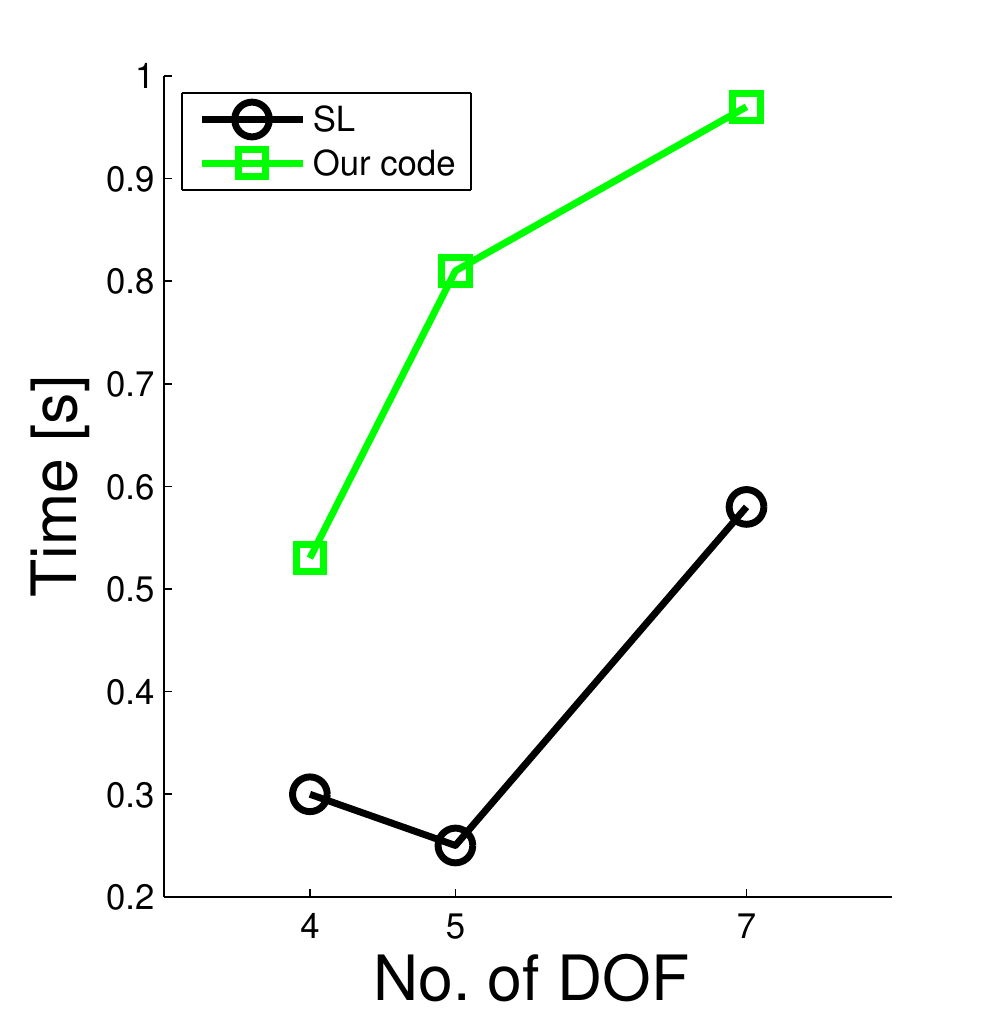}
	\caption[Performance comparisons of inverse dynamics algorithms.]
	{Performance comparison of two implementations of the Newton--Euler
	algorithm for inverse dynamics. The plot shows the cumulative
	execution time for $10^5$ calls of the function
	$\bm{\tau} = f(\ddot{\bm{q}},~\bm{q},\dot{\bm{q}})$ as a function of
	the number of degrees of freedom of the model (tests
	executed on a Intel(R) Core(TM)2 Duo CPU, P8700 @ 2.53GHz).}
	\label{fig:speedplot}
\end{figure}
%%%%%%%%%%%%
The graph in Figure \ref{fig:speedplot} shows how the algorithms scale as a
function of the number of \dofs, and shows at the same time a speed comparison.
We used a four-\dof robot (a leg of \hyq attached to a vertical slider), a
five-\dof robot with revolute and prismatic joints and finally a seven-\dof
model obtained adding a two link branch to the previous robot.
As expected, \SL provides the fastest function, but our implementation based
on Eigen is not that much slower. For some reasons, probably related to the
different optimization applicable to different models, \SL is slightly faster
with five \dofs compared to four \dofs.\\
It is important to note that we did not apply so much hand-crafted
optimization, leaving this job to the library and the compiler, but we got
already good results. Our
generator basically unrolls loops, uses a sparse vector for the motion
subspace matrix, and then quite literally maps the steps of the algorithm
into the appropriate algebra operations. Thus there is room for further
optimization, like precomputing some operations (\eg algebra which involves
constants such as $\tmxX{J}{p}$). Thanks to the overloaded operators
and clear identifiers built from the names provided in
the \dsl, the resulting code is human-readable, unlike the code generated
by \SL.\\
Even though this code does not have to be directly maintained -- as opposed to
what is behind the generator, \ie the model and the template -- having it
readable is quite desirable. The user can more easily inspect it, and spot
errors in the generation process.

All the code has been compiled with the same flags, and functions have been
statically linked into the executable. The program
measures \textsc{cpu} time by calling the library function
\textcode{std::clock()}.
%%%%%%%%%%%%%%%%%%%%%%%%%%%%%%%%%%%%%%%%%%%%%%%%%%%%%%%%%%%%%%%%%%%%%%%%%%%%%%%%
\section{RELATED WORK}\label{sec:related}
Software engineering \emph{for} robotics has only recently become an explicit
research area (especially if considering the age of the two disciplines), as
shown for example by the birth of a new journal \cite{brugali:2010:joser}.

In this context, the model--driven paradigm is recognized to be an effective
approach for the design of software. In \cite{steck:2010:qos}, the authors
point out the importance of resource awareness in robotics applications; they
describe a development process and a meta--model for robotics systems that
are focused on the non-functional properties of the components.\\
The techniques of meta--modeling and domain specific languages are exploited
in \cite{reckhaus:2010:prog_env} to design a programming environment independent
of the target robot, to facilitate the specification and reuse of control
programs.\\
In \cite{klotzbucher:2010:lua}, the authors present an execution environment
based on
the scripting language Lua, to support the implementation of internal \dsls for
modeling expressive state machines for robot coordination. The work focuses
particularly on dynamic memory management issues, not to violate real time
constraints during the interpretation (execution) of the state machines.

An example of the use of a \dsl in robotics, as a consequence of the need to find
higher abstractions to drive software development, is presented in
\cite{bordignon:2010:model_kin}, which targets the specific field of modular
robots. Here the authors give an
extensive description of a domain specific language for modeling the kinematics of
individual robot modules and their possible interconnections, which is exploited
to generate code for both the Webots simulator and a custom platform for the
execution of real experiments. In the same context,
\cite{schultz:2007:dsl_reconfigurable} presents a high level language built around
the concepts of roles to facilitate the programming of controllers for the modular
robot ATRON, independently of its physical configuration. While sharing the
approach of model based generation and the focus on kinematics, our work targets
the different domain of robots with linear or branched structure composed by rigid
links (such as manipulators or legged machines); it focuses on the generation
of efficient dynamics algorithms applicable in different components of a
software framework for robots.

\textsc{Simulink} \cite{matlab} is a well known tool in engineering which
supports the simulation of a broad class of dynamical systems, and can also
generate \matlab or C code.
However, \textsc{Simulink} is very general and thus not so convenient for
very specific needs like customized code generation of particular algorithms
as the ones by Featherstone \cite{featherstone:2008:rbda}.\\
Similar comments apply for instance for Modelica, a multi-domain,
object-oriented modelling language used also in industry \cite{web:modelica}.
Its models basically contain the system equations, which then need to
be transformed into executable code or into a form suitable for a simulation
engine.\\
Being so general purpose, these tools are likely to incur some unnecessary
overhead in terms of learning, usage and required tools, if one wants to get
similar results as with the \dsl; the \dsl infrastructure is more lightweight
and designed explicitly with the requirements of a real time controller code
for a real robot in mind.
%%%%%%%%%%%%%%%%%%%%%%%%%%%%%%%%%%%%%%%%%%%%%%%%%%%%%%%%%%%%%%%%%%%%%%%%%%%%%%%
%
\addtolength{\textheight}{-3cm}   % This command serves to balance the column lengths
\section{CONCLUSIONS AND FUTURE WORKS}\label{sec:end}
In this paper we have proposed a Domain Specific Language for the specification
of kinematics and dynamics parameters of robots consisting of rigid links.
The \dsl is based on a domain model that captures the minimum amount
of information required to specify the physics of the system. By using this
information it is possible to generate executable code as for instance
rigid body dynamics algorithms; such code is efficient and compatible with
real time constraints at high frequencies, \eg in low level control loops.
This approach allows researchers to
quickly set up new simulations or controllers, without having to deal
manually with critical and delicate parts of code.

This work aims at contributing to the field of model--based development
for robotics; on one side robotics research requires a lot of experimental and
exploratory activities and on the other side exhibits
many recurring issues and common problems that should be solved by
principled, general approaches. Our work aims at addressing these recurring
issues and thus freeing resources for the required exploratory sides
of robotics research.\\
However we stress that our work is still at a preliminary stage, and many
aspects could be improved. A natural development is to investigate other
targets for the code generation, addressing for instance algorithms for
floating base robots, and forward dynamics. As an additional example,
much of the
infrastructure for more advanced control schemes, \eg operational space
control \cite{khatib:1987:op_space, sentis:2007:phd} could be generated.
This includes
transformation, Jacobian and projection matrices for specific points on
the kinematic tree.\\
Other improvements of the \dsl itself include extending the validation of
documents with checks of semantic constraints (\eg a link cannot
be the child of more than one other link) or the usage in the documents of
labels defined externally.

The model described in section \ref{sec:uml} has been developed mainly as a
reference for the design of the \dsl, and could be refined and extended as
well. A minor improvement would be including data about the joints range of
motion, which is not relevant for dynamics algorithms but it is definitely
part of a kinematic description. Kinematic loops should be addressed
explicitly; additional classes like \textcode{Chain} and \textcode{Tree}
might be added.\\
In the paper we have already referred to the class diagram of Figure
\ref{fig:uml_model} as a meta--model.
As a final remark, we observe that it could equivalently be considered as
a simple
domain model, that is, a description which drives the design of software
representations of the important elements for a problem.
Joints and links (or legs, for example) are the subject of a variety of
tasks which involve different components of the robot software,
as for instance the low level position control or the planning of foot
trajectory in a humanoid. Therefore, finding proper representations in
computer code of these objects -- and of all the other relevant aspects --
is itself an important issue in the software for robotics.
%%%%%%%%%%%%%%%%%%%%%%%%%%%%%%%%%%%%%%%%%%%%%%%%%%%%%%%%%%%%%%%%%%%%%%%%%%%%%%%%
\section{ACKNOWLEDGMENTS}
This research has been funded by the Fondazione Istituto Italiano di Tecnologia.\\
We would like to thank Frank Buchli at Zuehlke Engineering for helpful
discussions about the use and implementation of domain specific languages.
%%%%%%%%%%%%%%%%%%%%%%%%%%%%%%%%%%%%%%%%%%%%%%%%%%%%%%%%%%%%%%%%%%%%%%%%%%%%%%%%
% BIBLIOGRAPHY
\bibliographystyle{plainurl}
\bibliography{biblio}
\end{document}